\documentclass[10pt,twocolumn,letterpaper]{article}

\usepackage{iccv}
\usepackage{times}
\usepackage{epsfig}
\usepackage{graphicx}
\usepackage{amsmath}
\usepackage{amssymb}
\usepackage{subfigure}
\usepackage[american]{babel}
\usepackage{color}
\usepackage{pdfpages}
\usepackage[pagebackref=true,breaklinks=true,letterpaper=true,colorlinks,bookmarks=false]{hyperref}

\iccvfinalcopy % *** Uncomment this line for the final submission
\def\iccvPaperID{1387} % *** Enter the ICCV Paper ID here

\ificcvfinal\fi
\begin{document}

%%%%%%%%% TITLE
\title{Deep Concept-wise Temporal Convolutional Networks for Action Localization}

\author{Xin Li$^1$, Tianwei Lin$^1$, Xiao Liu$^1$, Chuang Gan$^2$, Wangmeng Zuo$^3$, Chao Li$^1$,\\ Xiang Long$^1$, Dongliang He$^1$, Fu Li$^1$, Shilei Wen$^1$\\
Department of Computer Vision Technology (VIS), Baidu Inc.$^1$\\
MIT-Watson AI Lab.$^2$ Harbin Institute of Technology.$^3$\\
%{\tt\small {lixin41,lintianwei01,liuxiao12,lichao40,longxiang,hedongliang01,lifu,wenshilei}@baidu.com}
}

\maketitle
%\thispagestyle{empty}

%%%%%%%%% ABSTRACT
\begin{abstract}
Existing action localization approaches adopt shallow temporal convolutional networks (\ie, TCN) on 1D feature map extracted from video frames.
In this paper, we empirically find that stacking more conventional temporal convolution layers actually deteriorates action classification performance, possibly ascribing to that all channels of 1D feature map, which generally are highly abstract and can be regarded as latent concepts, are excessively recombined in temporal convolution.
%To address this issue, we introduce a novel concept-wise temporal convolutional network (C-TCN) as an alternative to TCN for training deeper action localization networks.
%
To address this issue, we introduce a novel concept-wise temporal convolution (CTC) layer as an alternative to conventional temporal convolution layer for training deeper action localization networks.
Instead of recombining latent concepts, CTC layer deploys a number of temporal filters to each concept separately with shared filter parameters across concepts. Thus can capture common temporal patterns of different concepts and significantly enrich representation ability.
%making performance consistently improved by increasing network depth.
%
Via stacking CTC layers, we proposed a deep concept-wise temporal convolutional network (C-TCN), which boosts the state-of-the-art action localization performance on THUMOS'14 from 42.8 to 52.1 in terms of mAP(\%), achieving a relative improvement of 21.7\%. Favorable result is also obtained on ActivityNet.

\end{abstract}

%%%%%%%%% BODY TEXT
\section{Introduction}
Video content analysis  methods have been extensively adopted in many real-world applications, such as entertainment, visual surveillance, and robotics.
A critical problem in video content analysis is action recognition for understanding human behavior and intent, but action recognition of manually trimmed video clip is unrealistic in practice, where the start and end time of each action instance are usually not annotated for real-world videos.
%
%Action recognition plays a critical role in understanding human behavior and intent, and can be extensively adopted in many real-world applications, such as entertainment, visual surveillance, and robotics.
%
%However, action recognition of manually trimmed video clip is unrealistic in practice, where the start and end time of each action instance are usually not annotated for real-world videos.
%
To address this issue, intensive studies have been gradually given to temporal action localization, and considerable progress has been made in the detection and localization of action instances in untrimmed video~\cite{chao2018rethinking,lin2017single,yuan2017temporal}.

%The advances in smart phones, communication, and computing are stimulating the transit of traditional products to their video gene.
%Technologies in video content analysis are in high demand to boost the user experience.
%However, the performance of temporal action localization is the bottleneck of video content analysis.

\begin{figure}[t]
\begin{center}
\begin{minipage}[b]{1.0\linewidth}
  \centering
  \centerline{\includegraphics[width=8.5cm]{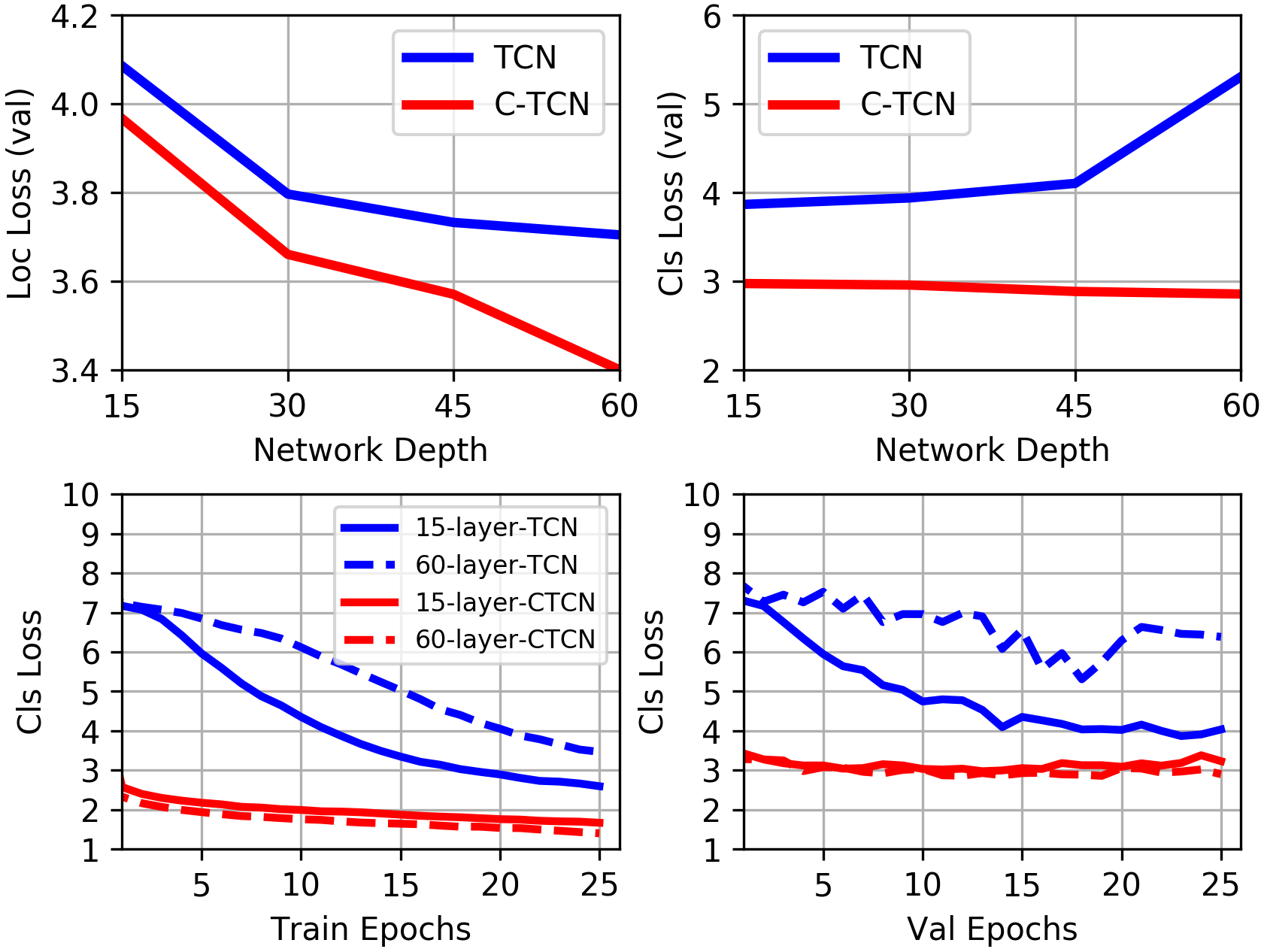}}
  \medskip
\end{minipage}
\end{center}
   \caption{Localization losses (top left) and classification losses (top right) of TCNs and C-TCNs with different network depths on the ActivityNet validation set. Along with the increase of network depth, the classification loss of TCN becomes unexpectedly larger. The training and testing curves (bottom left/bottom right) indicate that stacking more conventional temporal convolution layers deteriorates category modeling rather than improving it. }
\label{fig:pre}
\end{figure}

Temporal action localization and spatial object detection share similar task paradigms.
In particular, object detection aims to find objects in a 2D image in the form of the bounding box positions and object categories.
Analogously,  the goal of temporal action localization is to detect action instances in a 1D sequence of frames and output the temporal boundaries as well as action categories.
%Object detection is a fundamental computer vision task and has achieved remarkable progresses thanks to the development of deep learning in recent years.
Motivated by such similarity, one can transfer the existing 2D spatial convolutional network (CNN) into 1D temporal convolutional network (TCN) for temporal action localization~\cite{lea2016temporal, lea2017temporal, chao2018rethinking, lin2017single, lin2018bsn}.
For example, TAL-Net~\cite{chao2018rethinking} and SSAD~\cite{lin2017single} can be regarded as the temporal version of Faster-RCNN~\cite{ren2015detection} and SSD~\cite{liu2016detection}, respectively.

However, the transfer from 2D spatial CNN to 1D TCN is not always straightforward and proper architecture design of TCN is still required.
Concretely, increasing the depth of backbone network~\cite{he2016resnet} generally is helpful in improving object detection performance~\cite{ren2015detection, liu2016detection, lin2017fpn, cornernet}.
In contrast, simply stacking more layers in conventional TCN may even cause performance degradation in temporal action localization.
%
%Architecture design is one of the most critical components for object detection.
%The backbone networks of state-of-the-art object detection methods~\cite{ren2015detection, liu2016detection, lin2017fpn, cornernet} are usually very deep and gain performance improvements from stacking multiple residual blocks~\cite{he2016resnet}.
%In contrast, most previous 1D-TCNs~\cite{lea2016temporal, lea2017temporal, chao2018rethinking, lin2017single, lin2018bsn} are relatively shallow with less than 10 layers in their backbones.
%A natural question to ask is: \emph{could 1D-TCNs benefit from increased network depth?}
%
To illustrate this point, we train four TCNs on ActivityNet~\cite{Heilbron2015activitynet} under the same architecture except that the network depths are different (\ie, 15, 30, 45, 60).
All models are optimally tuned on the training set and tested on the validation set.
As shown in  Fig.~\ref{fig:pre} (a) and (b), the localization performance can be consistently improved along with the increase of network depth, while the classification performance begins to drop when the network depth is higher than 15.
Fig.~\ref{fig:pre}(c) and (d) further 
%illustrate the training and testing curves of TCNs with network depths of 15 and 60, 
indicate that the degradation of classification performance cannot be simply explained by over-fitting to training data.

Why performance of conventional TCN degraded with deeper network?
To begin with, the input feature sequence $\mathbf{F}$ of TCN can be denoted as $c \times t$, where $t$ and $c$ are number of snippets and number of feature channels separately.
Benefited from deep appearance and motion feature extraction, channels of $\mathbf{F}$ generally are highly abstract and can be regarded as disentangled {\bf concepts}.
In a conventional temporal convolution layer, each temporal filter is operated on all $c$ concepts and output the weighted combination.
This operation contains two possible drawbacks: (1) the concepts may be excessively recombined, making concepts in deep feature maps are not discriminate enough for action classification; (2) each concept is generated by one temporal 1D convolution filter. If we term the number of filters used in TCN for generating a concept as "Potential", it will be 1 for ordinary temporal 1D.  Small Potential can suffer from inferior capability.  
These drawbacks may explain the performance degradation of deeper TCN, and make deep TCN not a proper choice to boost performance of action localization.
A straightforward way to reduce the recombination of concepts is group convolution, which divides the input $\mathbf{F}$ into $c$ groups and each group is convolved with its own set of $k$ temporal filters.
Thus, the shape of the convolved feature map $\mathbf{F}^{\prime}$ will be $ (kc) \times t$, where the filters of different groups have independent parameters.
However, our experiments in Sec. 4.2 show that naively utilizing group convolutions provides little benefit in improving the performance of TCN.

Considering that the extracted feature sequence can be viewed as the response sequence of different concepts, we intuitively expect \emph{common temporal patterns of different concepts can be captured}.
To achieve this, we propose a novel {\bf concept-wise temporal convolution (CTC) layer}, which deploys a number of temporal filters to each concept separately but shares the filter parameters across concepts.
Thus, CTC layer can (1) reduce the concepts recombination, (2) enrich concept context via expanding a concept from one potential to multiple potentials, and (3) capture common temporal patterns via sharing the filter parameters across concepts.
We implement CTC layer in a natural way, via expanding the 1D snippet-level representation into a 2D map, whose two dimensions are potential and concept respectively.
In such representation, CTC can be directly achieved and the shape of the convolved feature map will be $ k \times c \times t$. 
%Thus, the snippet-level representation is transformed from 1D to 2D, making it more structured.
%A C-TCN is obtained by stacking multiple CTC layers.
Based on stacked CTC layers, we proposed the {\bf Concept-wise Temporal Convolution Network (C-TCN)}, which is an anchor-based temporal action localization network with deep backbone.

%To address the limitations of conventional TCN, we further present a kind of concept-wise temporal convolution (CTC) as well as a C-TCN model by stacking multiple concept-wise convolution layers.
%We treat each channel of feature map $\mathbf{F}$ as a concept, and different concepts are not recombined but a group of temporal convolutions are deployed to each concept individually. In particular,

%In particular, we treat each channel of feature map $\mathbf{F}$ as a conceptual dimension.

%Then, CTC is introduced by employing $K$ 2D convolution filters with kernel size of $1 \times 3$  to each concept.
%Thus, the size of convolved feature map $\mathbf{F}^{\prime}_C$ will be $K \times H_c \times T$, and the snippet-level representation in hidden layers will be 2D.
%C-TCN can be formed by stacking multiple convolution layers.
%In contrast to conventional TCN, CTC is effective in enriching representation ability, and can achieve better tradeoff between efficiency and performance due to concept-wise convolution.

%Extensive experiments have been conducted on ActivityNet and THUMOS'14.
%The results show that C-TCN outperforms TCN and can be consistently improved by increasing network depth.
%On both datasets C-TCN can achieve state-of-the-art performance and performs favorably against the competing methods.

In summary, our work has three main contributions:
\begin{itemize}
\item We systematically analyze  why performance of conventional TCN degraded with deeper network, and reveal that the key solutions are reducing concept combination and capturing common temporal patterns of different concepts to enrich representation ability.

\item We propose an novel and effective Concept-wise Temporal Convolution (CTC) layer that allows training deeper model for action localization, and propose a deep Concept-wise Temporal Convolution Network (C-TCN) based on CTC layer.

\item Extensive experiments show that C-TCN can outperforms TCN consistently with increasing network depth, and can achieve state-of-the-art performance on both ActivityNet-1.3 and THUMOS-14 datasets.

\end{itemize}

\section{Related Work}
\subsection{Action Recognition}
Action recognition is a fundamental task in video content analysis.
The goal of action recognition is to classify a trimmed video with only one action instance.
Early approaches proposed many hand-crafted features as spatial-temporal representations~\cite{hog3d, sift3d, mbh, idt}.
Thanks to the progress made by deep convolution neural network, ConvNet based methods achieved superior accuracy to conventional hand-crafted features.
Pre-trained ConvNets are used in~\cite{LRCN, shuttleNet, tle, netvlad, actionvlad, attentioncluster} to extract convolutional features from video frames,
which are encoded into a global representation for classification.
Recent methods~\cite{simonyan2014action, karpathy2014action, two-stream-stresnet, wang2016action, tran2015action, carreira2017action, qiu2017action, ARN, nonlocal, s3d} also explored to design network architectures to exploit the spatial-temporal patterns of actions.

\begin{figure*}[t]
\setlength{\abovecaptionskip}{-0.3cm} %缩小caption和图像之间的距离
\setlength{\belowcaptionskip}{-0.3cm} %缩小caption和下方文字的距离
\begin{center}
\begin{minipage}[b]{1.0\linewidth}
  \centering
  \centerline{\includegraphics[width=16cm]{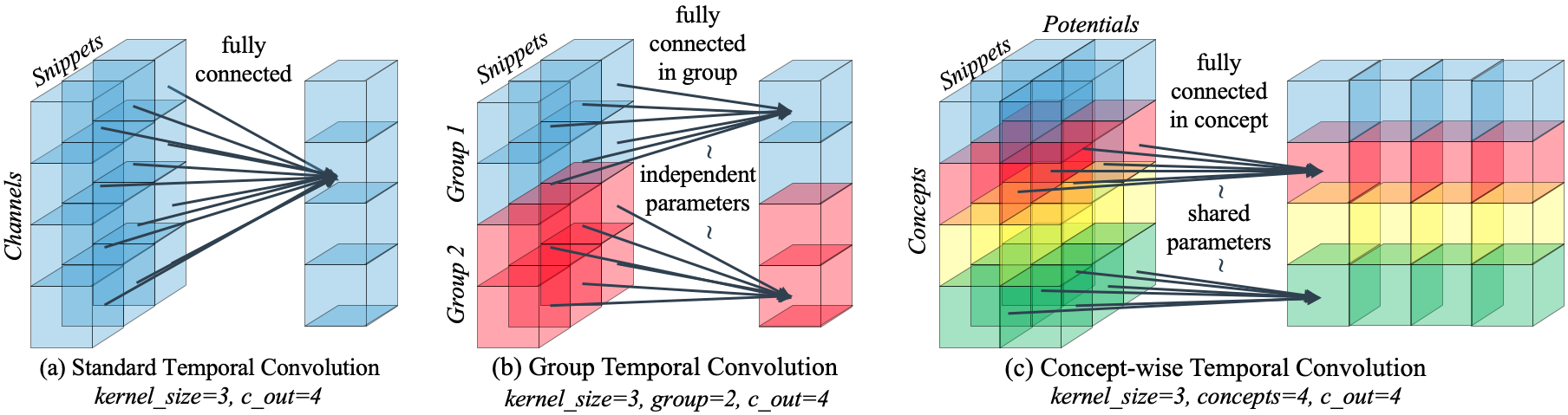}}
  \medskip
\end{minipage}
\end{center}
\vspace{-0.5cm}
   \caption{  Illustration of standard, group and concept-wise temporal convolutions. (a) Standard temporal convolution recombines all concepts. (b) Group temporal convolution recombines concepts in each group separately with independent parameters. (c) Concept-wise temporal convolution recombines potentials in each concept separately with shared parameters. Thus the discrimination of snippet-level representation is best preserved and the common temporal patterns can be captured.}

\label{fig:conv}
\vspace{-0.3cm}
\end{figure*}

\subsection{Temporal Action Localization}
Early works apply hand-crafted features and sliding windows~\cite{tang2013localization, yuan2016localization} for temporal action localization.
\emph{E.g.}, Yuan et al.~\cite{yuan2016localization} used SVM to encode the iDT features~\cite{idt} at each position and each resolution, followed by sliding window to localize the actions.
In order to combine multiple hand-crafted features, Tang et al.~\cite{tang2013localization} introduced a hierarchical graph method for feature selection and learned the structure of the graph using score-based structure learning.
Similarly, Heilbron et al.~\cite{heilbron2016dict} learned sparse dictionaries from multiple features to represent and retrieve activity proposals.

Snippet-level deep features such as two-stream CNN~\cite{simonyan2014action}, C3D~\cite{tran2015action} and I3D~\cite{carreira2017action} have achieved both higher efficiency and better performance than hand-crafted features.
LSTM is used to generate action proposals~\cite{escorcia2016temporal, yeung2016rl} and detection segments~\cite{ma2016learning, kaustend},
while 1D-temporal convolutional architectures~\cite{lea2016temporal, lea2017temporal, lin2017single} show better performance than LSTM when modeling long range temporal structure of actions.
Inspired by state-of-the-art region-based object detectors~\cite{girshick2014detection, girshick2015detection, ren2015detection, liu2016detection, lin2017fpn, cornernet},
two-stage~\cite{shou2017cdc, chao2018rethinking, shou2016temporal, heilbron2017scc, gao2017turn, dai2017temporal, gao2017cascaded, wang2017untrimmednets, zhao2017temporal} and one-stage~\cite{lin2017single, lin2017short} action detectors were proposed for the temporal domain.
Xu et al.~\cite{xu2017r} encoded video streams with a trainable C3D network to generate proposals and classified the selected proposals into specific activities, such that the entire model is trained end-to-end from video frames.
Some other works firstly generate snippet-level action scores and then use the labels for contextual reasoning.
A few of them~\cite{xiong2017tag, lin2018bsn, yuan2017temporal} aimed to find the start and end boundaries of action proposals, while~\cite{richard2016action, hou2017real} jointly optimized the appearance, temporal structure and action duration for action detection with structure modeling.

\section{Proposed Method}
\subsection{Snippet-level Feature Coding}
Following ~\cite{lea2016temporal, lea2017temporal, lin2017single}, our model is built upon deep appearance and motion features extracted from raw video frames.
We first uniformly divide the video into several small consecutive snippets and then extract visual features within each snippet.
In our implementation, the output scores of the ``pool5'' layer of a pre-trained two-stream model ~\cite{simonyan2014action} are concatenated as the snippet-level representation.
Thus, the shape of the visual feature sequence $\mathbf{F}$ is $c \times t$, where $c$ is the dimension of the snippet-level representation, and $t$ is the number of snippets.

\subsection{Concept-wise Temporal Convolution Layer}
The channels of $\mathbf{F}$ can be regarded as disentangled concepts.
Motivated by our analysis of conventional TCN, we expect the concept-wise temporal convolution (CTC) layer to meet three demands:
(i) context of each concept should be enriched;
(ii) each concept should be convolved separately to avoid the recombination of concepts;
(iii) fiter parameters should be shared across concepts to capture common temporal patterns.

In CTC layer, we view the snippet-level feature as an $1\times c$ tensor
such that the shape of $\mathbf{F}$ is $1\times c \times t$.
Considering the reshaped $\mathbf{F}$ as an 1-channel map whose height and width are $c$ and $t$,
we can use $k$ $1\times 3$ (3 could be other size) filters to convolve the map in temporal dimension and obtain a new convolved feature map $\mathbf{F}^{\prime}$ whose shape is $k\times c \times t$.
As shown in Fig.~\ref{fig:conv}, rectangular filters are utilized such that the concepts are not mixed together.
Meanwhile, the filter parameters are shared by all concepts.
%due to that different concepts may share similar common temporal patterns.
Thus, CTC layer expands the original $\mathbf{concept}\times \mathbf{snippet}$ feature sequence to the new $\mathbf{potential} \times \mathbf{concept}\times \mathbf{snippet}$ feature map.
The potential number of the input data is 1, and the potential number of a hidden layer decided by the filter kernel number of the previous CTC layer.
%Fig.~\ref{fig:conv} illustrates the procedure.
The CTC layer can also be achieved with group convolution and weight sharing.
But our implementation is more natural and concise.

%The conceptual feature map can also be viewed as a $H_c$-channel 1D feature sequence whose length is $T$.
%Existing 1D-TCNs~\cite{lea2016temporal, lea2017temporal, chao2018rethinking, lin2017single, lin2018bsn}
%perform 1D convolutions along its temporal dimension.
%As a result, the conceptual information is tangled together after each convolutional layer and deeper 1D-TCN suffers more seriously.

%Note that the conceptual feature map can be regarded as a gray-scale image whose height is $H_c$ and width is $T$,
%and we can directly use it as the input of a spatial 2D ConvNet.
%The problem is if we use typical 2D convolution with square kernels, information of adjacent conceptual dimensions is mixed for no reason.
%In order to address this issue, we propose concept-wise convolutions which conduct 2D convolution operations with $1\times K$ kernels.
%Similar with 1D temporal convolutions, the concept-wise convolutions are also performed along the temporal dimension.
%In contrast with 1D temporal convolutions, the information of each conceptual dimension is modeled separately but not tangled together.
%We also used $1\times S$ stride in concept-wise convolutions, such that the conceptual information is guaranteed to be preserved.

\begin{figure*}[t]
\begin{center}
\begin{minipage}[b]{1.0\linewidth}
  \centering
  \centerline{\includegraphics[width=16cm]{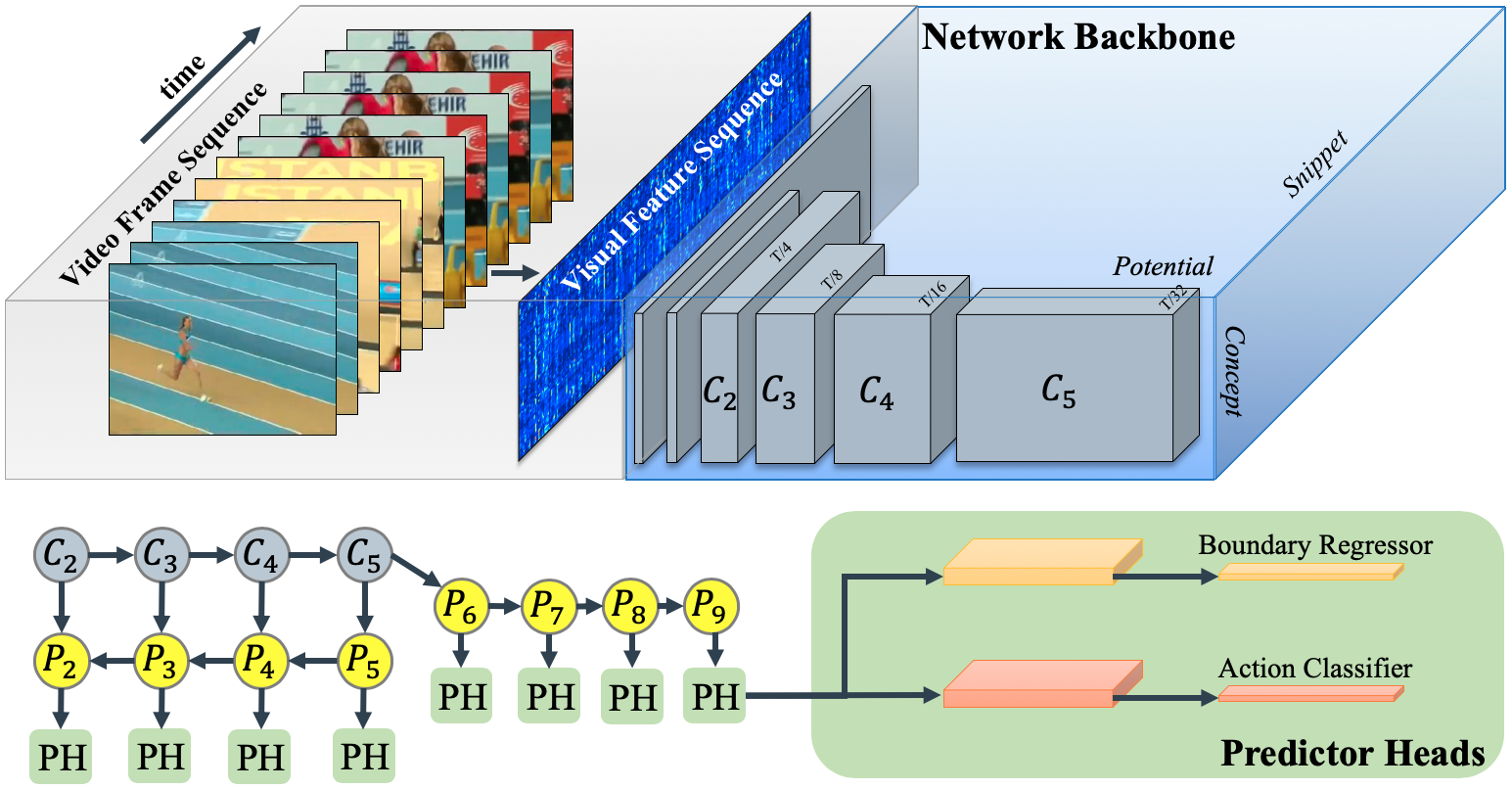}}
  \medskip
\end{minipage}
\end{center}
\vspace{-0.5cm}
   \caption{The backbone and feature pyramid of our C-TCN.
             The backbone is a ResNet with 4 stages of CTC residual blocks, from $C_2$ to $C_5$.
             We then build feature pyramid of 8 scales (only four are shown here), from $P_2$ to $P_9$, on the top of the residual blocks.
             Two predictor heads for temporal segment classifier and regressor are then adopted for each scale of feature pyramid.}
\label{fig:flowchart}
\vspace{-0.3cm}
\end{figure*}

\subsection{Concept-wise TCN}
Based on CTC layer, it is available to build a deep Concept-wise Convolution Network (C-TCN).
%transfer the existing object detection frameworks of 2D spatial CNN to C-TCNs.
In this paper, we transfer existing object detection framework - feature pyramid network (FPN)~\cite{lin2017fpn} to build C-TCN as an anchor-based action localization network.
%In this paper, we exploit the object detection framework of feature pyramid network (FPN)~\cite{lin2017fpn} to build C-TCN.
However, the transfer is not restricted to specific framework.
The backbone and FPN adopted in our C-TCN is illustrated in Fig.~\ref{fig:flowchart}.

\noindent\textbf{Backbone.} We build a feature pyramid on the top of the ResNet architecture~\cite{he2016resnet}.
Our feature pyramid consists of 8 temporal scales, from $P_2$ to $P_9$, where the resolution of $P_l$ is $1/2^l$ of the input.
The resolution of the finest scale $P_2$ is $1/4$ of the original input, while the resolution of the coarsest scale $P_9$ is $1/512$ of the original input.
Specifically, the snippet number of the input feature sequence is 512, and we use two $1\times 7$ CTC layers with $1\times2$ stride to decrease its temporal length to 128.
The feature map is then passed to a ResNet with 4 stages of CTC residual blocks.
The activation outputs of the four residual stages are denoted from $C_2$ to $C_5$.
$P_6$ is computed via a $1\times 3$ CTC layer with $1\times2$ stride on $C_5$.
$P_7$ to $P_9$ are obtained by applying ReLU followed by a $1\times 3$ CTC layer with $1\times2$ stride on the previous scale correspondingly.
We then compute $P_2$ to $P_5$ based on $C_2$ to $C_5$ using top-down and lateral connections as in~\cite{lin2017fpn}.
The architecture and the parameter setting of our CTC are tabulated in supplementary material.

\noindent\textbf{Predictor Head.}
We use two predictor heads for temporal segment classifier and regressor separately.
Each predictor head has a hidden layer followed by a prediction layer.
The hidden layer is a CTC layer that decreases the potential number.
The prediction layer is a $c\times1$ convolutional layer that combines information from all concepts.
It produces a $(A+1)M$-channel $2^{9-l}$-length score map for $P_l$,
where $A+1$ is the number of action categories plus one background class, $l$ is the pyramid scale and $M$ is the number of anchor segments.
Hence, the temporal segment classifier produces classification scores for $M$ anchor segments at each temporal location and scale of pyramid.
Similarly, the prediction layer of the regressor produces a $2M$-channel $2^{9-l}$-length output for $P_l$.
The location and size offsets are predicted for each anchor segment at each temporal location and scale of pyramid.

\noindent\textbf{Default Size and Position.}
Different scales of pyramid have different basic segment sizes.
The basic segment size of $P_l$ is set as:
\begin{equation}
s_l = \frac{2^l}{512}T, \hspace{0.1in} l = 2\dots9,
\end{equation}
such that the coarsest scale $P_9$ has the basic segment size of $T$, which is the length of the input video.
For each scale of pyramid, we use $M$ anchor segments to adjust the basic segment size subtly.
Denote the default size of the $m^{th}$ anchor segment of $P_l$ as $s_{l,m}$, we have
\begin{equation}
s_{l,m} = \frac{2}{3}s_l + \frac{2(m-1)}{3M}s_l, \hspace{0.1in}m = 1\dots M,
\end{equation}
for the pyramid scale from $P_2$ to $P_8$.
It means that $P_l$ occupies a length range from $\frac{2}{3}s_l$ to $\frac{4}{3}s_l$,
and the range is divided into $M$ equal portions associated to different anchor sizes.
The coarsest scale $P_9$ occupies the length range from $\frac{2}{3}s_9$ to $s_9$,
because the length of an action can not be longer than the length of the video.

Regarding that $P_l$ has $2^{9-l}$ cells, and each one is associated with a default center location of action segment.
The default center location of the $j^{th}$ cell in $P_l$ is
\begin{equation}
b_{l,j} = \frac{(2j-1)}{2^{10-l}}T.
\end{equation}

\noindent\textbf{Classification and Regression.}
For simplicity, we denote $prop_{i}^j$ as a temporal proposal of video $V_i$ with index $j$,
and $\hat{w}_{i,j}$ and $\hat{c}_{i,j}$ are the default length and center temporal location of the proposal obtained by (1), (2) and (3)
according to its pyramid scale, the cell index and the anchor index.

For action classification, we reshape the output of the classifier in each cell to a $(A+1)\times M$ matrix,
where each column of the matrix ($o^0, o^1, \dots, o^A$) is related to the scores of $A+1$ categories,
and each row is related to an anchor.
A softmax layer is adopted to transform the scores to class confidences, and we denote
\begin{equation}
\hat{y}_{i,j}^k = \frac{\exp{o_{i,j}^k}}{\sum_{k'}{\exp(o_{i,j}^{k'})}}
\end{equation}
as the $k^{th}$ class confidence of $prop_{i}^j$.

The output of the regressor in each cell is reshaped to a $2\times M$ matrix,
where the first element in each column is the length offset and the second element in each column is the location offset.
The regressed length of $prop_{i}^j$ is $\exp(\hat{\beta}_{i,j})\hat{w}_{i,j}$
and the regressed center temporal location of $prop_{i}^j$ is $\hat{c}_{i,j} + \hat{w}_{i,j}\hat{\gamma}_{i,j}$,
where $\hat{\beta}_{i,j}$ and $\hat{\gamma}_{i,j}$ are length and location offsets of $prop_{i}^j$ correspondingly.

\noindent\textbf{Matching Strategy and Training Objective.}
We need to match the temporal proposals to ground-truth actions to train the network.
Given $\hat{w}_{i,j}$ and $\hat{c}_{i,j}$, we can calculate the temporal Intersection-over-Union (tIoU) between $prop_{i}^j$ and the ground-truth actions in $V_i$.
Each temporal proposal is matched to the ground-truth action with the largest tIoU, and is regarded as a positive sample if its largest tIoU is larger than $0.5$,
otherwise  a negative sample.

Our training objective function combines classification loss and localization loss in a multi-task framework.
The classification loss of a positive temporal proposal is a cross-entropy form:
\begin{equation}
L_{cla}^{pos}(prop_{i}^j) = -\log{(\hat{y}_{i,j}^{y_{i,\zeta(i,j)}})},
\end{equation}
where $\zeta(i,j)$ is the index of the matched ground-truth of $prop_{i}^j$.
Similarly, the classification loss of a negative temporal proposal is
\begin{equation}
L_{cla}^{neg}(prop_{i}^j) = -\log{(\hat{y}_{i,j}^0)}.
\end{equation}

The localization loss is only applied to positive samples.
Inspired by SSD~\cite{liu2016detection} and Faster-RCNN~\cite{ren2015detection}, we regress the parametric offsets instead of the start and end boundaries:
\begin{eqnarray}
L_{loc}(prop_{i}^j) = {\rm smooth}_{L1} (\hat{\beta}_{i,j} - \beta_{i,j})   \nonumber \\
       +{\rm smooth}_{L1} (\hat{\gamma}_{i,j} - \gamma_{i,j})
\end{eqnarray}
where $\beta_{i,j} = \log({\frac{w_{i,\zeta(i,j)}}{\hat{w}_{i,j}}})$
and
$\gamma_{i,j} = (c_{i,\zeta(i,j)} - \hat{c}_{i,j}) / \hat{w}_{i,j}.$

We thus have the overall training objective as follows:
\begin{eqnarray}
L(\tau, \theta) = \sum_i \left( \sum_{j\in \Psi_{i}}L_{cla}^{neg}(prop_{i}^j)+  \hspace{0.2in}\right.\nonumber  \\
             \left .  \sum_{j\in \Omega_{i}}\left(L_{cla}^{pos}(prop_{i}^j) + L_{loc}(prop_{i}^j)\right) \right),
\end{eqnarray}
where $\Omega_{i}$ and $\Psi_i$ denote the positive sample set and negative sample set of $V_i$ respectively, and $\theta$ indicates learnable parameters.

\noindent\textbf{Hard Negative Mining and Data Augmentation.}
Following the spirit of modern object detection approaches~\cite{ren2015detection, liu2016detection, lin2017fpn, cornernet},
For hard negative mining, we apply hard negative mining and data augmentation during training.
Instead of using all the negative samples, we pick the negative samples with higher classification loss,
and keep the ratio between the negatives and positives samples as $3:1$.

Data augmentation is useful for preventing overfitting, but is not adopted in temporal action localization methods yet.
In this work, we propose two data augmentation strategies: temporal {\bf random move} and {\bf random crop}.
In random move, we first remove all action segments from the video, and concatenate the left background parts to a new video. We then randomly insert action segments to the concatenated video.
In random crop, we randomly crop a temporal segment from the video and resize it the size of the original video.
At least 50\% of a ground-truth action is guaranteed to be left after random crop.

\section{Experimental Results}
To evaluate the proposed approach, we conduct experiments on two video action localization datasets, \emph{i.e.} ActivityNet-1.3~\cite{Heilbron2015activitynet} and THUMOS'14~\cite{jiang204thumos}.
The impact of different components of our algorithm is investigated by ablation studies.
The comparison between C-TCN with other state-of-the-art methods are also reported.
\subsection{Experimental Settings }
\noindent\textbf{Dataset.} ActivityNet-1.3 consists of more than 648 hours of untrimmed sequences from a total of ~20K videos.
It contains 200 different daily activities such as ``walking the dog'', ``long jump'' and ``vacuuming floor''.
The numbers of videos for training, validation and testing are 10,024, 4,926, and 5,044.
We report results on its validation set because the labels of testing set are not released.
THUMOS'14 has 20 action classes with 200 untrimmed videos in validation set and 213 videos in testing set.
We use its validation set to train our model and report results on its testing set.

\noindent\textbf{Evaluation metrics.} We follow conventional evaluation metrics of  action localization task:  mean Average Precision (mAP) with different tIoU thresholds. On ActivityNet-1.3, mean of all mAP values computed with tIoU thresholds between 0.5 and 0.95 (inclusive) with a step size of 0.05 is used. On THUMOS'14, mAPs with tIoU thresholds {0.1, 0.2, 0.3, 0.4, 0.5} are adopted.
We use the AR-AN (average recall with average number of proposals) curve to evaluate the quality of generated temporal proposals.

\noindent\textbf{Features.} To extract snippet-level video feature, we first sample frames from each video at 5fps on both ActivityNet-1.3 and THUMOS'14 datasets, and we then apply a TV-L1~\cite{tvl1} algorithm to get the optical flow of each frame.
For ActivityNet-1.3, two TSN~\cite{wang2016action} models with Senet152~\cite{senet} backbones are separately trained on RGB and stacked optical flow, and the two trained models are used to extract two-stream features.
For THUMOS'14, we use the TSN models pre-trained on ActivityNet-1.3 to extract two-stream features. An open source I3D model~\cite{carreira2017action} pre-trained on Kinetics is also employed to extract the I3D features.
We use an additional linear layer to reduce the dimension of snippet-level representation to 256.

\noindent\textbf{Implementation Details.}
In the training phase, we train the models using Stochastic Gradient Descent (SGD) with momentum of 0.9, weight decay of 0.0001 and a mini-batch size of 16.
Additional dropout layers with an ratio of 0.5 are added after the feature maps of all pyramid scales.
On THUMOS'14, we set the initial learning rate at 0.001 and reduce once with a ratio of 0.1 after 200 epochs.
On ActivityNet-1.3, we set the initial learning rate at 0.0005 and reduce once with a ratio of 0.1 after 20 epochs.
The model is trained from scratch and an early-stop strategy is also used to prevent model overfitting.
In the testing phase, we apply Soft-NMS for each action class separately and merge the outputs of all classes as the final results.

\subsection{Ablation Studies}
%Our ablation studies are based on the THUMOS'14 dataset, because experiments on ActivityNet are very time consuming.

\noindent\textbf{C-TCN vs. TCN.}
We trained C-TCNs and conventional TCNs with different numbers of layers on THUMOS'14 and the results are summarized in Table \ref{table:ctcn_tcn}.
As can be seen, C-TCNs can gain accuracy from increased depth but TCNs get degraded results when the layer number is larger than 15.
Furthermore, we see that C-TCNs perform better than TCNs consistently with the same network depth.
 In Table \ref{table:ctcn_tcn}, we also demonstrate the performance of TCN with different groups, which indicates that solely avoid concept recombination in TCN is not enough, but sharing filter parameters is necessary for boosting detection performance.

\begin{table}[t]
\caption{Comparison of C-TCN vs. conventional TCN on THUMOS'14 in terms of mAP (\%).  \emph{L} is the number of layers in network, and \emph{G} is the group number of TCN. }
\vspace{-0.4cm}
\begin{center}
\begin{tabular}{llllllll}
\hline
 & \emph{L} & \emph{G}  & 0.1   & 0.2   & 0.3   & 0.4   & 0.5   \\ \hline
TCN & 15 & 1 & 60.0 & 58.5 & 53.9 & 49.0 & 41.0 \\
TCN & 30 & 1 & 63.2 & 61.5 & 57.9 & 51.6 & 42.4 \\
TCN & 45 & 1 & 58.0 & 55.3 & 51.7 & 45.6 & 38.1 \\
TCN & 60 & 1 & 56.4 & 53.8 & 50.4 & 43.8 & 36.3 \\ \hline
TCN & 60 & 4 & 57.9 & 55.5 & 51.8 & 46.0 & 36.0 \\
TCN & 60 & 16 & 59.6 & 57.2 & 52.7 & 46.1 & 36.8 \\
TCN & 60 & 64 & 62.6 & 59.8 & 55.2 & 47.9 & 36.1 \\
TCN & 60 & 256 & 39.5 & 36.4 & 31.5 & 24.8 & 18.1 \\ \hline
C-TCN & 15 & - & 70.9 & 69.4 & 66.6 & 59.7 & 49.0 \\
C-TCN & 30 & - & 71.2 & 69.9 & 67.2 & 60.3 & 50.5 \\
C-TCN & 45 & - & 71.4 & 70.3 & 67.9 & 60.9 & 51.4 \\
C-TCN & 60 & - & \textbf{72.2} & \textbf{71.4} & \textbf{68.0} & \textbf{62.3} & \textbf{52.1} \\
\hline
\end{tabular}
\end{center}
\label{table:ctcn_tcn}
\end{table}

\noindent\textbf{Feature Setting.}
The comparison of performance between using TSN and I3D features are summarized in Table \ref{table:feature}.
As can be seen, the I3D features achieve better performance than the TSN features.
As expected, combining the RGB and flow features obtains much higher mAP than using a single modality,
demonstrating that the two modalities are heavily complementary for this task.
We also investigate the effect of different fusion methods and find early fusion achieving slight better results than late fusion.

\begin{table}[t]
\caption{Study of different feature settings on THUMOS'14 in terms of mAP(\%)@tIoU.}
\vspace{-0.3cm}
\begin{center}
\begin{tabular}{llllll}
\hline
tIoU                      & 0.1   & 0.2   & 0.3   & 0.4   & 0.5   \\ \hline
TSN RGB                   & 58.9 & 56.5 & 51.4 & 44.0 & 35.0 \\
TSN flow                  & 47.0 & 44.9 & 42.3 & 37.6 & 31.1 \\
TSN R. + F. late & 63.0     & 61.6     & 57.8     & 50.7     &  41.7     \\
TSN R. + F. early & 66.0 & 64.5 & 59.8 & 52.6 & 42.5 \\
\hline
I3D RGB                   & 60.0     & 58.1     & 54.6     & 47.2     & 37.6 \\
I3D flow                  & 67.4     & 66.5     & 63.5     & 57.3     & 49.7 \\
I3D R. + F. late & 70.7 & 69.3 & 66.4 & 61.2 & 51.5 \\
I3D R. + F. early & \textbf{72.2} & \textbf{71.4} & \textbf{68.0} & \textbf{62.3} & \textbf{52.1} \\
\hline
\end{tabular}
\end{center}
\label{table:feature}
\end{table}

\noindent\textbf{Anchor Setting.}
We try using different number of anchor segments in our experiments, and the results are summarized in Table \ref{table:anchor}.
One can see that using 7 anchors achieves the best performance with the tIoU thresholds of 0.2, 0.3, 0.4 and 0.5,
and using 11 anchors achieves the best performance with the tIoU threshold of 0.1.
Using more anchors not always obtains better results because more redundant detections are involved.
We also compare the performance of using the anchor setting of SSD~\cite{liu2016detection} with 6 anchors.
All its mAPs with different tIoUs are lower than our setting with 7 anchors.
We can conclude that our new setting is better than the SSD anchor setting for this task.

\begin{table}[t]
\caption{Study of different anchor settings on THUMOS'14 in terms of mAP(\%)@tIoU.}
\vspace{-0.3cm}
\begin{center}
\begin{tabular}{llllll}
\hline
tIoU               & 0.1   & 0.2   & 0.3   & 0.4   & 0.5   \\
\hline
SSD anchor setting & 71.5 & 69.9 & 64.9 & 58.7 & 47.4 \\
3 anchors          & 68.8 & 67.9 & 64.6 & 59.1 & 49.3 \\
5 anchors          & 70.1 & 68.9 & 65.0 & 58.4 & 47.7 \\
7 anchors          & 72.2 & \textbf{71.4} & \textbf{68.0} & \textbf{62.3} & \textbf{52.1} \\
9 anchors          & 69.2 & 67.8 & 64.7 & 58.4 & 48.9  \\
11 anchors         & \textbf{73.2} & 71.2 & 67.9 & 58.7 & 46.6 \\
13 anchors        & 68.4 & 66.7 & 63.7 & 57.2 & 47.2 \\
\hline
\end{tabular}
\end{center}
\label{table:anchor}
\end{table}

\noindent\textbf{Data Augmentation Setting.}
The comparison of different data augmentation settings is summarized in Table \ref{table:data_aug}.
It can be seen that data augmentations play critical roles in preventing model overfitting.
Without data augmentation, the model converges very fast and the mAP drops seriously.
We also find that random move and random crop are highly complementary.

\begin{table}[t]
\caption{Study of different data augmentation settings on THUMOS'14 in terms of mAP(\%)@tIoU.}
\vspace{-0.3cm}
\begin{center}
\begin{tabular}{llllll}
\hline
tIoU              & 0.1   & 0.2   & 0.3   & 0.4   & 0.5   \\
\hline
no random crop    & 68.6 & 67.2 & 63.9 & 56.2 & 45.1 \\
no random move    & 63.3 & 62.1 & 59.4 & 54.8 & 44.2 \\
no augmentation   & 60.2 & 58.2 & 54.3 & 47.7 & 37.4 \\
both augmentations & \textbf{72.2} & \textbf{71.4} & \textbf{68.0} & \textbf{62.3} & \textbf{52.1} \\
\hline
\end{tabular}
\end{center}
\label{table:data_aug}
\end{table}

\noindent\textbf{Network Architecture Settings.}
We also conduct ablation studies for other network architecture settings, and the results are summarized in Table \ref{table:network}.
Five different settings are conducted to demonstrate the superiorities of the components used in our method.
(1) With the regression step during testing removed, the mAP with the threshold of 0.5 drops from 52.1 to 41.4.
(2) With the kernel size from $1\times 3$ changed to $3\times3$, which means that we use square 2D convolutions instead of concept-wise convolutions,
the mAP drops to 45.4.
(3) With the stride size from $1\times 2$ changed to $2\times2$, which means that the conceptual dimension is downsampled at higher pyramid scale, the mAP drops to 39.9.
(4) With the pyramid scale number reduced from 8 to 7 and the finest scale removed,
the mAP drops to 44.5.
(5) With the top-down and lateral connections removed, the mAP drops to 47.7.
The results together prove that the network architecture of C-TCN has been optimally tuned.

\begin{table}[t]
\caption{The ablation studies of network architecture settings on THUMOS'14 in terms of mAP(\%) with tIoU 0.5.}
\begin{center}
\begin{tabular}{lcc}
\hline
Settings & mAP (\%)\\\hline
remove regression            &  41.4 \\
kernel size changed to square   & 45.4 \\
stride size changed to square   &  39.9\\
reduce scale number  &  44.5\\
remove top-down connections &  47.7\\
\hline
\end{tabular}
\end{center}
\label{table:network}
\end{table}

\subsection{Comparison with the State-of-the-arts}

\begin{figure}[h]
	\begin{center}
		\includegraphics[scale = 0.5]{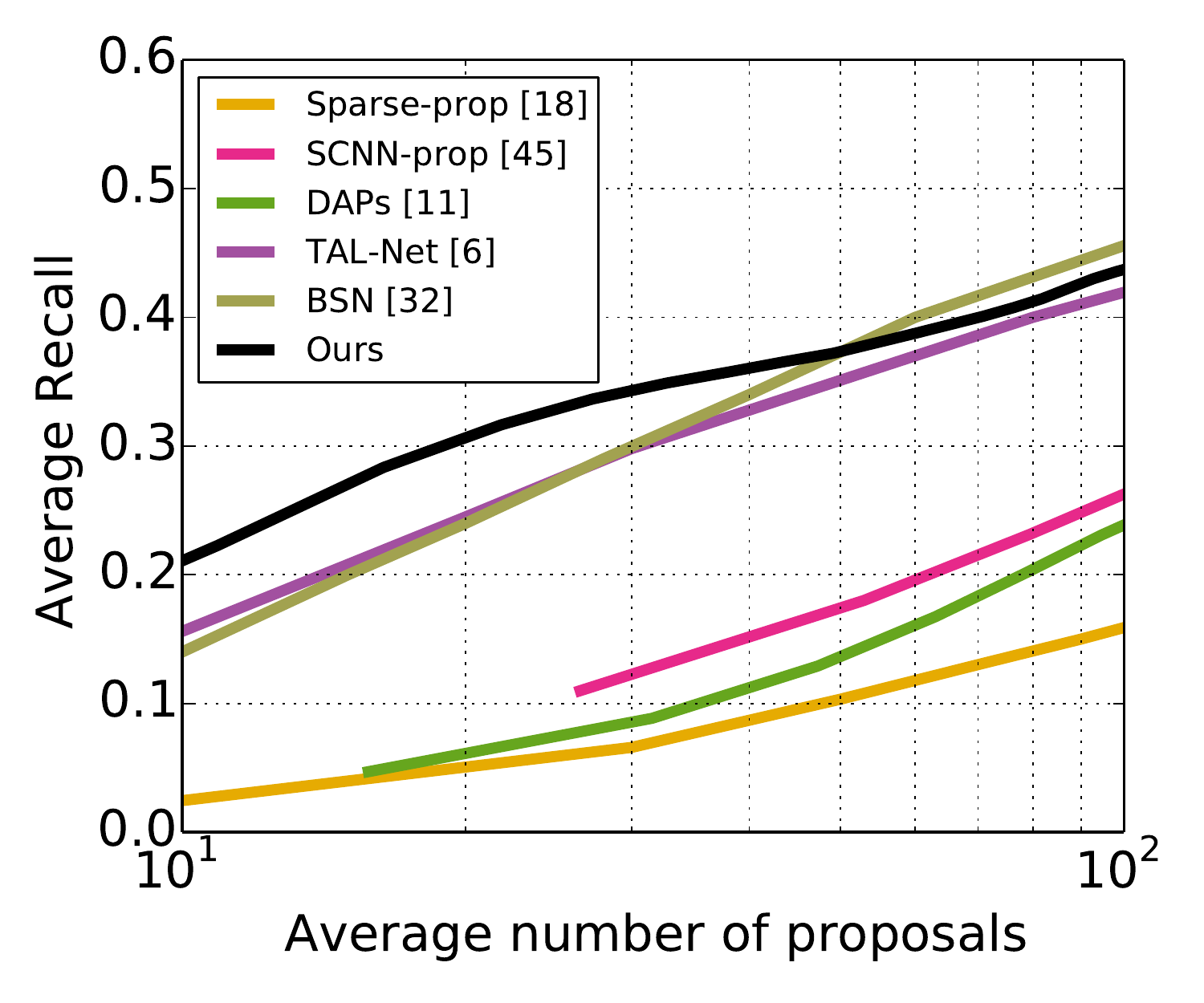}
	\end{center}
	\caption{Comparisons in the AR-AN (\%) on THUMOS'14.
             }
   	\label{fig:prop}
\end{figure}

\noindent\textbf{THUMOS'14.}
We compare C-TCN with other state-of-the-art methods on THUMOS'14 and summarize the results in Table~\ref{table:thumos}.
As our results are significantly better than others, we compare our method with others by evaluating the quality of generating action proposals for further analysis.
Figure~\ref{fig:prop} shows the AR-AN curves for action proposals.
We find that our method achieves higher AR than others in the low AN region, but is not the best when AN is higher than 40.
On the one hand, the curve in low AN region is more closely related to the mAP metrics used in action localization.
On the other hand, boundary information is potentially highly complementary with our anchor-based method.
\begin{table}[t]
\caption{Action localization mAP(\%) on THUMOS'14.}
\begin{center}
\begin{tabular}{llllll}
\hline
tIoU               & 0.1  & 0.2  & 0.3  & 0.4  & 0.5  \\ \hline
Karaman et al.~\cite{karaman2014thumos}\hspace{-0.2in}    & 4.6  & 3.4  & 2.4  & 1.4  & 0.9  \\
Oneata et al.~\cite{oneata2014thumos}      & 36.6 & 33.6 & 27.0 & 20.8 & 14.4 \\
Wang et al.~\cite{wang2014thumos}       & 18.2 & 17.0 & 14.0 & 11.7 & 8.3  \\
Richard et al~\cite{richard2016action}   & 39.7 & 35.7 & 30.0 & 23.2 & 15.2 \\
Shou et al.~\cite{shou2016temporal}        & 47.7 & 43.5 & 36.3 & 28.7 & 19.0 \\
Yeung et al.~\cite{yeung2016localization}        & 48.9 & 44.0 & 36.0 & 26.4 & 17.1 \\
Yuan et al.~\cite{yuan2016localization}        & 51.4 & 42.6 & 33.6 & 26.1 & 18.8 \\
Shou et al.~\cite{shou2017cdc}        & -    & -    & 37.8 & -    & 23.0 \\
Buch et al.~\cite{buch2017action}        & -    & -    & 45.7 & -    & 29.2 \\
Gao et al.~\cite{gao2017cascaded}         & 60.1 & 56.7 & 50.1 & 41.3 & 31.0 \\
Dai et al.~\cite{dai2017temporal}         & -    & -    & -    & 33.3 & 25.6 \\
Gao et al.~\cite{gao2017turn}         & 54.0 & 50.9 & 44.1 & 34.9 & 25.6 \\
Xu et al.~\cite{xu2017r}          & 54.5 & 51.5 & 44.8 & 35.6 & 28.9 \\
Zhao et al.~\cite{zhao2017temporal}       & 66.0 & 59.4 & 51.0 & 41.0 & 29.8 \\
Chao et al.~\cite{chao2018rethinking} & 59.8 & 57.1 & 53.2 & 48.5 & 42.8 \\
Lin et al.~\cite{lin2018bsn} & -    & -    & 53.5 & 45.0 & 36.9 \\ \hline
Ours               & \textbf{72.2} & \textbf{71.4} & \textbf{68.0} & \textbf{62.3} & \textbf{52.1} \\
\hline
\end{tabular}
\end{center}
\label{table:thumos}
\end{table}

\begin{figure*}[t]
	\begin{center}
		\includegraphics[scale = 0.62]{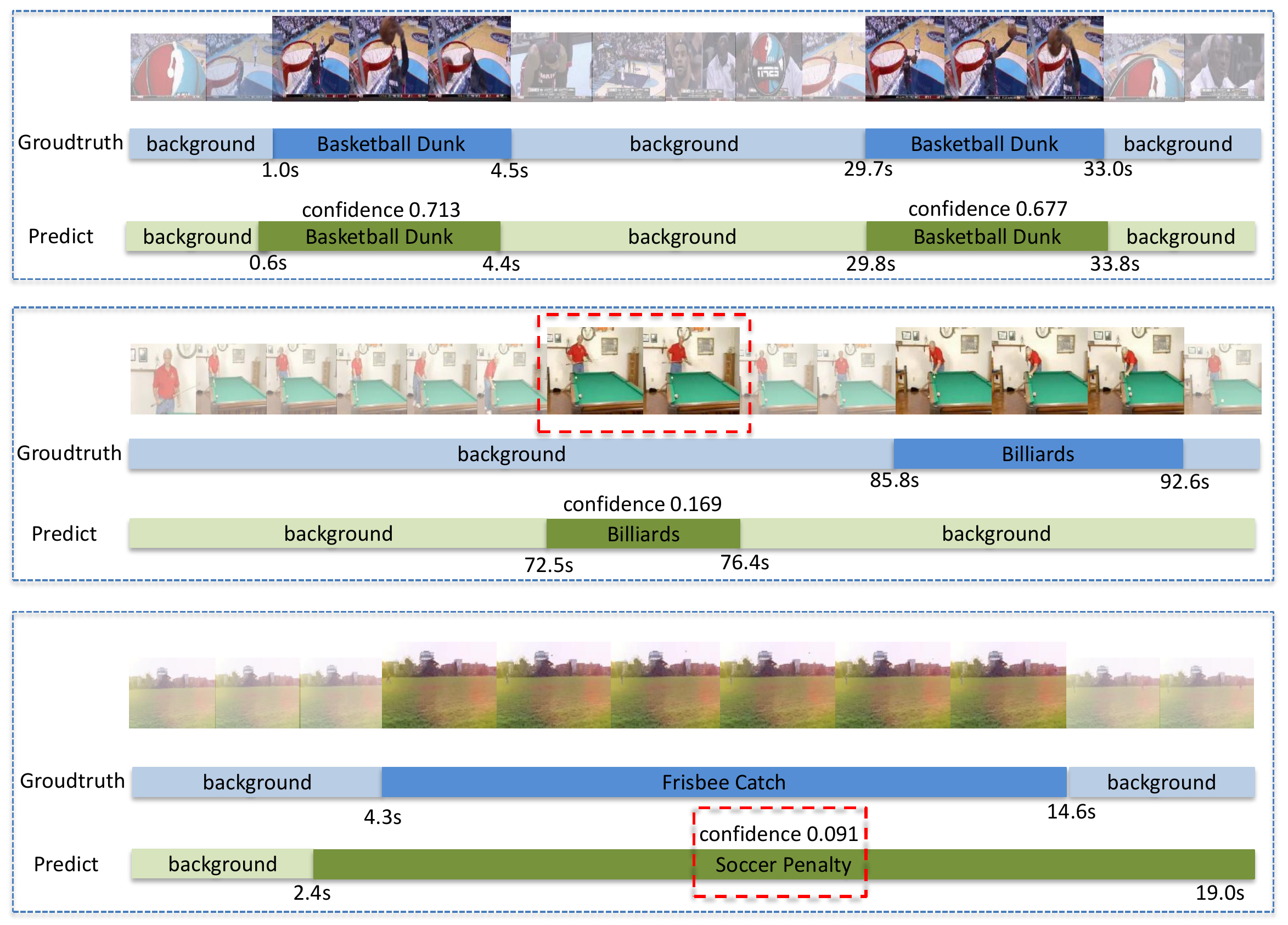}
	\end{center}
	\caption{Qualitative results on THUMOS'14 with a good case on the top row and two bad cases on the middle and the bottom rows. }
   	\label{fig:example}
\end{figure*}

\noindent\textbf{ActivityNet-1.3.}
Our comparisons on ActivityNet-1.3 are summarized in Table~\ref{table:anet}.
We can see that the proposed method again achieves the best results among all competing methods.
BSN~\cite{lin2018bsn} has a higher mAP than our C-TCN with the threshold of $0.95$, however, it uses additional video classification results of~\cite{cuhk2017cls} which are fused by multiple models,
while our model only uses a two-stream model.

\begin{table}[th]
\caption{Action localization mAP(\%) on ActivityNet v1.3 (val).
* means additional video-level classification results are fused.
}
\begin{center}
\begin{tabular}{lllll}
\hline
tIoU               & 0.5   & 0.75  & 0.95 & Ave. \\
\hline
Singh et al.~\cite{singh2016ant} & 34.4  & -     & -    & -       \\
Wang et al.~\cite{wang2016ant}      & 43.6  & -     & -    & -       \\
Heilbron et al.~\cite{heilbron2017scc}    & 40.0 & 17.9 & 4.7 & 21.7   \\
Shou et al.~\cite{shou2017cdc}        & 45.3  & 26.0  & 0.2  & 23.8    \\
Dai et al.~\cite{dai2017temporal}         & 36.4  & 21.2  & 3.9  & -       \\
Xu et al.~\cite{xu2017r}          & 26.8 & -     & -    & 12.7   \\
Chao et al.~\cite{chao2018rethinking} & 38.2  & 18.2  & 1.3  & 20.2    \\
Lin et al.~\cite{lin2017single}*        & 44.3 & 29.6 & 7.0 & 29.1   \\
Lin et al.~\cite{lin2018bsn}*         & 46.4 & 29.9 & \textbf{8.0} & 30.0   \\
\hline
Ours               & \textbf{47.6}      & \textbf{31.9}      & 6.2     & \textbf{31.1}   \\
\hline
\end{tabular}
\end{center}
\label{table:anet}
\end{table}

\subsection{Qualitative Results}
Qualitative results on THUMOS'14 with a good case on the top row and two bad cases on the middle and the bottom rows are shown in Figure~\ref{fig:example}.
In the first bad case, a ground-truth billiard action is undetected and a wipe cue stick action is wrongly detected as the billiard action.
In the second bad case, our algorithm successfully localize the period that the action happens, but misclassified the frisbee catch action to soccer penalty.

\section{Conclusion}
In this paper, we studied the challenging task of temporal action localization in video.
Delving into the performance degradation along with the increase of network depth, we empirically reveal that degradation of classification performance may ascribe to that all concepts are recombined in temporal convolution.
In this work, we present a novel concept-wise temporal convolutional network (C-TCN) to improve temporal action localization.
The C-TCN can transfer the popular architecture from spatial 2D CNN and achieves accuracy gains consistently by increasing network depth. Experimental results show that C-TCN brings substantial improvements over the conventional TCN and achieves state-of-the-art performance in ActivityNet and THUMOS'14.
In future we will exploit C-TCNs for video classification.
Codes are available at  \href{https://github.com/PaddlePaddle/models/tree/develop/PaddleCV/PaddleVideo}{PaddleVideo}.

{\small
\bibliographystyle{ieee}
\bibliography{egbib}
}

\newpage

\includepdf[pages=1,pagecommand={}]{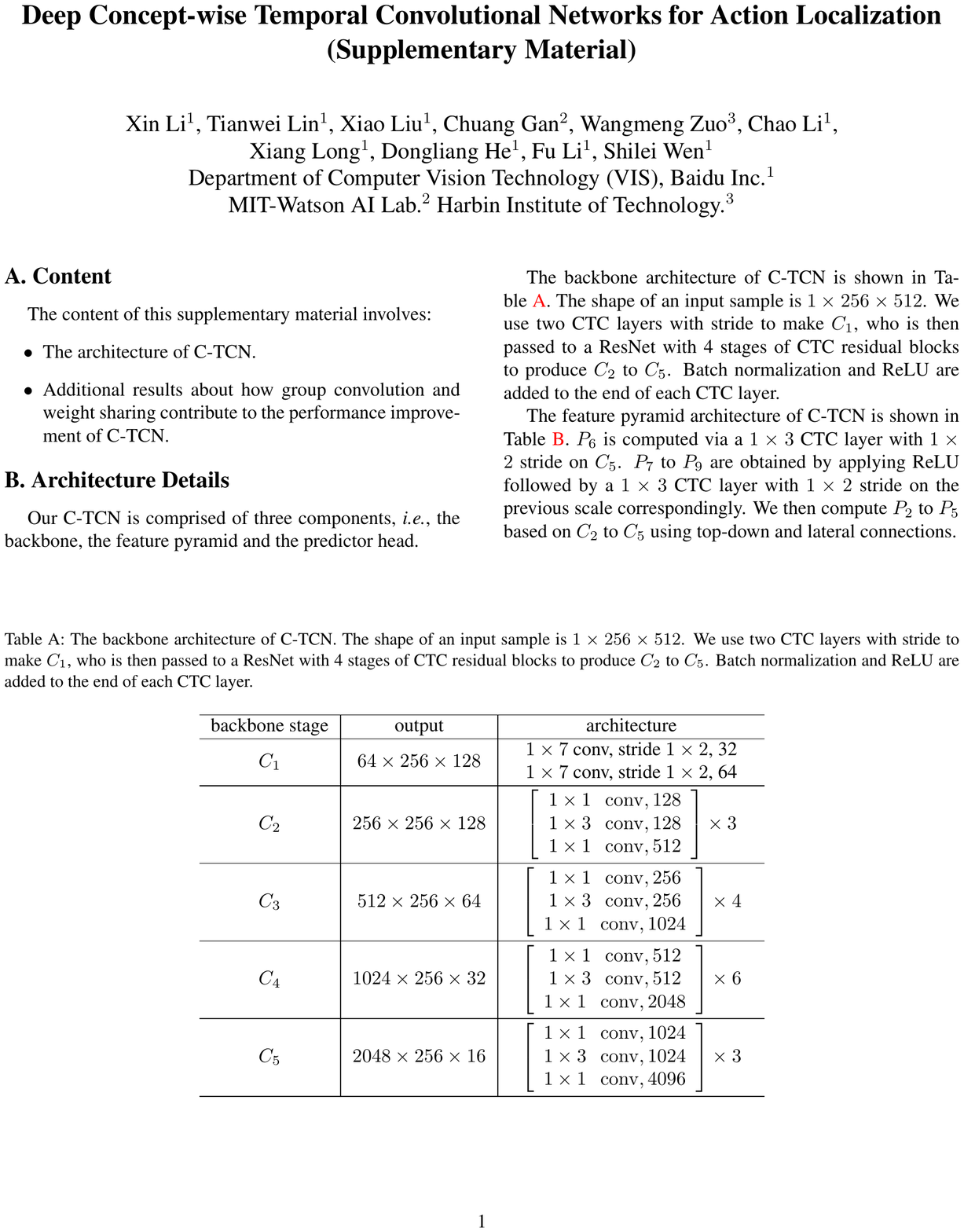}
\includepdf[pages=2,pagecommand={}]{CTCN_SM.pdf}
\includepdf[pages=3,pagecommand={}]{CTCN_SM.pdf}

\end{document}